\PassOptionsToPackage{numbers,sort&compress}{natbib}
\documentclass[sigconf,nonacm]{acmart}

\usepackage[utf8]{inputenc}
\usepackage[T1]{fontenc}
\usepackage{microtype}
\usepackage{booktabs}
\usepackage{tabularx}
\usepackage{multirow}
\usepackage{float}
\usepackage{amsmath,amssymb}
\usepackage{xcolor}
\usepackage{tikz}
\usepackage{url}
\usetikzlibrary{arrows.meta,positioning,calc,fit}

\setcopyright{none}
\hypersetup{
  hidelinks,
  pdftitle={When Guardrails Look Effective: Construct Validity Failures in LLM Agent Commerce Evaluation},
  pdfauthor={Peiying Zhu; Sidi Chang}
}

\definecolor{oldred}{HTML}{B94C4C}
\definecolor{controlblue}{HTML}{2474A6}
\definecolor{validgreen}{HTML}{3E8E64}
\definecolor{warnamber}{HTML}{C8861A}
\definecolor{softgray}{HTML}{ECEFF1}
\definecolor{ink}{HTML}{263238}

\newcommand{\none}{\textsc{None}}
\newcommand{\info}{\textsc{Info}}
\newcommand{\conduct}{\textsc{Conduct}}
\newcommand{\both}{\textsc{Both}}
\newcommand{\invalid}{\textsc{Invalid}}
\newcommand{\inconclusive}{\textsc{Inconclusive}}

\begin{document}
\acmshorttitle{When Guardrails Look Effective}
\acmshortauthors{Zhu \& Chang}
\pagestyle{acmstyle}
\thispagestyle{firstpage}

\twocolumn[
\begin{@twocolumnfalse}
\begin{center}
{\LARGE\bfseries When Guardrails Look Effective: Construct Validity Failures in LLM Agent Commerce Evaluation\par}
\vspace{1.0em}

\begin{tabular}{@{}c@{\hspace{2em}}c@{}}
{\large Peiying Zhu\textsuperscript{*}} & {\large Sidi Chang\textsuperscript{*\,$\dagger$}} \tabularnewline
{\small peiying@blossomai.co} & {\small schang@blossomai.co} \tabularnewline
{\small Blossom AI} & {\small Blossom AI Labs} \tabularnewline
{\small San Francisco, CA, USA} & {\small Tokyo, Japan}
\end{tabular}
\vspace{1.2em}
\end{center}

\noindent\parbox{\textwidth}{
\small
\noindent\textbf{Abstract}\\[3pt]
Interactive simulations are increasingly used to evaluate policies for markets populated by language-model agents. Their outputs can look economic---prices, profits, consumer surplus, and welfare---even when the simulation does not instantiate the economic behavior named in the claim. We audit this risk in a multi-turn buyer--seller testbed for configurable hotel transactions. An initial implementation reported welfare gains from two marketplace guardrails of $+87.4$, $+35.0$, and $+28.8$ across a Qwen2.5 1.5B--14B ladder. That implementation also gave guarded and unguarded agents different offer schemas and choice procedures. Holding the schema and buyer chooser fixed changes the same paired contrasts to $+7.2$, $-13.9$, and $+23.8$. The four largest 14B single-generation effects averaged $+229$; after three generations per profile-condition, they averaged $+37.6$ (95\% bootstrap interval $[-34.2,109.3]$), while generation residuals account for 49.9\% of the variation in this post-hoc probe. A seller-incentive manipulation check is non-monotone: explicitly increasing profit pressure produces less profit than the default seller prompt. Scripted positive controls show why this matters. A profit-maximizing seller already attains first-best welfare, so guardrails mostly redistribute and reduce welfare; guardrails create welfare only when the seller is explicitly programmed to force inefficient bundles. We contribute a construct-validity contract for agent-market evaluation that separates incentive validity, protocol isolation, stochastic stability, and welfare accounting, and returns \invalid{} or \inconclusive{} before allowing a substantive policy claim. Applied to our own case, the original estimate is \invalid{} under protocol isolation, while the controlled study remains \inconclusive{} under incentive validity and stochastic stability. The case study does not establish that guardrails are ineffective; it establishes that their apparent value is unidentified until the simulated agents and protocol pass these checks.
}

\vspace{1.5em}
\end{@twocolumnfalse}
]

\renewcommand{\thefootnote}{\fnsymbol{footnote}}
\footnotetext[1]{Both authors contributed equally to this research.}
\footnotetext[2]{Corresponding author.}
\renewcommand{\thefootnote}{\arabic{footnote}}

\section{Introduction}

Agent-to-agent (A2A) commerce is becoming a concrete evaluation target. Buyer agents search and negotiate; seller agents configure offers and prices; platforms may filter messages or constrain offer formats. Recent environments make it possible to study such systems before deployment \cite{bansal2025magentic,xia2024bargaining}. The attraction is clear: assign hidden values and costs, let agents interact, and score completion, surplus, or welfare against ground truth.

There is a measurement trap inside this workflow. Calling an LLM a ``self-interested seller'' does not establish that it maximizes profit. Calling a prompt edit a ``guardrail'' does not establish that a measured effect comes from the rule rather than from a friendlier output schema or decision scaffold. Treating one stochastic dialogue per buyer profile as an experimental observation does not identify generation variance. Finally, a higher buyer surplus does not imply higher welfare: if a transaction still occurs, a lower price often transfers surplus from seller to buyer without creating any.

These are construct-validity problems, not cosmetic limitations. Construct validity asks whether an operationalization actually instantiates and measures the theoretical object named in a claim \cite{cronbach1955construct,jacobs2021measurement,bean2025measuring}. In an agent-market simulation, the relevant object is not merely a fluent transcript. It is a role, incentive, information set, action space, and outcome mapping. A simulation can be internally consistent yet fail to represent the strategic seller or welfare intervention it claims to evaluate.

We make this failure visible through a forensic case study. We began with a plausible A2A hypothesis: two thin platform guardrails---one blocking buyer information useful for personalized extraction and one preventing optional components from being forced into a mandatory base---could improve transaction outcomes. The original testbed produced a clean positive result. We then ran a control that removed an implementation asymmetry, repeated the most favorable profiles, tested whether seller prompts moved profit-seeking behavior, and introduced scripted seller policies as positive controls. Each step weakened the original headline while sharpening the evaluation lesson.

Our contributions are:
\begin{enumerate}
  \item We document a \emph{scaffold sensitivity}: an apparent guardrail effect changes sharply, including a sign reversal at 3B, when guarded and unguarded cells share one offer schema and one buyer choice rule.
  \item We quantify \emph{single-generation instability} without pseudo-replication. Replications are averaged within profile-condition, and profiles---not generation rows---are the paired unit. An exact profile-level sign-flip placebo finds the selected effect compatible with label exchangeability ($p=0.50$).
  \item We expose an \emph{incentive-validity gap}: prompt-defined seller roles do not respond monotonically to a stronger profit instruction. Scripted positive controls show that the welfare effect of guardrails changes sign with the assumed seller technology.
  \item We propose a compact evaluation contract and a three-way decision: substantive interpretation only after the contract passes; \invalid{} for a known construct violation; \inconclusive{} when precision or coverage is insufficient. Table~\ref{tab:verdict} applies the contract to this study, and the accompanying artifact provides a reusable nine-item checklist.
\end{enumerate}

This is a bounded methods result. We do not conclude that A2A guardrails fail, that the Qwen family is non-strategic in general, or that hotel negotiation represents every market. We conclude that policy claims from LLM market simulations require evidence that the intended economic constructs are present and that the treatment is isolated from its scaffold.

\section{Related Work}

\noindent\emph{LLM agents and economic simulation.}
LLMs have been proposed as simulated economic agents with assigned preferences and endowments \cite{horton2023homo}. AgentBench evaluates interactive decision-making across environments \cite{liu2023agentbench}; bargaining benchmarks formalize buyer and seller gains \cite{xia2024bargaining}; and Magentic Marketplace studies multi-agent markets with consumer and service agents \cite{bansal2025magentic}. Recent equilibrium-referenced supply-chain experiments show that prompt parameters and model-provider identity can dominate surplus division \cite{liang2026negotiate}, while LLM pricing agents can exhibit collusive behavior that changes under seemingly innocuous prompt wording \cite{fish2024collusion}. These works make role and incentive assumptions increasingly consequential.

\noindent\emph{Evaluation as measurement.}
Construct validity concerns the relation between an intended construct and its operationalization \cite{cronbach1955construct}. Measurement-modeling work emphasizes that observable metrics inherit assumptions about unobservable constructs \cite{jacobs2021measurement}; a recent review finds that such assumptions are often weakly articulated in LLM benchmarks \cite{bean2025measuring}. HELM similarly argues for multi-metric, scenario-aware evaluation rather than a single aggregate score \cite{liang2022helm}. Prompt-equivalent evaluations can vary substantially \cite{cao2024worst}; single-sample evaluation ignores stochastic generations \cite{wadi2025montecarlo}; and expensive agent evaluations often omit the repeated runs needed for error bars \cite{kapoor2025agents}. We specialize this lens to interactive economic simulations, where the environment partly creates the behavior later attributed to an agent or policy.

\noindent\emph{Trace and diagnostic validity.}
Prior trace-based work shows that scalar task performance can remain stable while hidden-state discipline fails, and that aggregate diagnostics can misrank policy repairs \cite{zhu2026trace,zhu2026aggregate}. Our setting differs in both construct and mechanism: we study multi-turn buyer--seller dialogue, stochastic role enactment, treatment scaffolds, and welfare accounting. The connection is methodological: outcomes do not identify whether an interactive system followed the behavioral assumptions needed to interpret them.

\section{Setting and Evaluation Contract}

\subsection{Configurable transaction testbed}

Each buyer profile $\theta$ has hidden component values, hard constraints, a willingness-to-pay cap, urgency and an outside option. A seller can offer a mandatory base set $S$ at price $p$ and separately priced optional additions. Components include cancellation, quiet room, breakfast, view, high floor, and late checkout. The buyer accepts the utility-maximizing feasible offer if it weakly beats the assigned outside option. Hotels are a worked configurable-goods instance, not the claimed universe of application.

All outcome metrics use assigned ground truth rather than values stated by the agents. For an accepted set $S$,
\begin{align}
  \mathrm{BS}_{\theta}(S,p) &= V_{\theta}(S)-p-o_{\theta},\\
  \mathrm{SP}_{\theta}(S,p) &= p-C(S),\\
  W_{\theta}(S) &= \mathrm{BS}+\mathrm{SP}=V_{\theta}(S)-C(S)-o_{\theta}.
\end{align}
All three are zero when no transaction occurs. This identity is central: price discrimination alone moves surplus between buyer and seller; it changes welfare only if it changes completion or composition. First-best welfare is the maximum of zero and welfare from the hard constraints plus each optional component whose private value exceeds its cost. We report welfare as a fraction of this profile-specific first best. Match quality credits only components in that welfare-maximizing set.

The platform has two binary rules. The information rule blocks questions and responses about budget, WTP, urgency, and outside options while preserving need-discovery questions. The conduct rule restricts the mandatory base to essential components and turns other components into declineable add-ons. Their $2\times2$ crossing yields \none{}, \info{}, \conduct{}, and \both{}.

\subsection{A validity contract before a policy claim}

We define four predicates that should be evaluated before interpreting $\Delta W$ as the effect of a marketplace policy:

\begin{description}
  \item[\normalfont C1: Incentive validity.] The seller measurably responds to its stated objective. Strengthening profit incentives should change profit-seeking behavior in the predicted direction, or the role must be described behaviorally rather than strategically.
  \item[\normalfont C2: Protocol isolation.] Treatment cells share message schema, decision rules, horizon, and parsing. Only the intended information and conduct constraints vary.
  \item[\normalfont C3: Stochastic stability.] The profile-level treatment effect exceeds generation noise at the intended decision threshold. Repeated generations are nested within profile-condition and are not counted as independent buyers.
  \item[\normalfont C4: Accounting completeness.] Completion, buyer surplus, seller profit, and welfare are reported together from ground truth. A distributional transfer is not labeled welfare creation.
\end{description}

The decision rule is deliberately asymmetric. A known violation of C1 or C2 makes the corresponding causal interpretation \invalid{}. For C1, however, an underpowered or non-monotone manipulation check is not proof that the intended incentive is absent; it leaves the incentive interpretation \inconclusive{}. Failure to resolve sampling or coverage under C3 is likewise \inconclusive{}, not ``no effect.'' Only after C1--C4 pass is a positive, negative, or practically null policy conclusion licensed.

\noindent\emph{Identifiability boundary.}
Let $z(\tau)$ be a binary trace property such as ``the seller used private WTP to set price,'' and let an evaluator observe only $\phi(\tau)$. If two traces $\tau_1,\tau_2$ satisfy $\phi(\tau_1)=\phi(\tau_2)$ but $z(\tau_1)\ne z(\tau_2)$, no downstream function of $\phi(\tau)$ can exactly recover $z$. Adding more aggregate metrics computed from the same abstraction cannot repair the missing instrumentation. This elementary observation motivates retaining prompts, visible and hidden transcripts, parsed offers, profile types, and costs rather than only scalar outcomes.

\section{Audit Design}

\subsection{Models, profiles, and protocol}

We run 4-bit Qwen2.5-Instruct models at 1.5B, 3B, and 14B parameters \cite{yang2024qwen25}. In each capability cell the same model instantiates buyer and seller. Conversations have two seller-question rounds followed by a structured offer; temperature is 0.2. We use the same 30 held-out synthetic profiles in every model and policy cell. A separately optimized fixed menu is descriptive context, not a treatment baseline for the validity audit, because it received distribution-level optimization that the zero-shot agents did not.

The synthetic profile generator is intentionally visible as a limitation. In the full 60-profile evaluation panel, 43 outside-option utilities are negative under $o_{\theta}=150-\text{outside price}-\text{quality penalty}$. Negative utility is mathematically coherent but makes acceptance easier and may not represent a target deployment population. We therefore avoid general market-level welfare claims from this panel.

\subsection{Four audit experiments}

\noindent\emph{E1: Scaffold control.}
The first implementation gave the unguarded seller a single-bundle \texttt{room\_attributes/price} schema and a whole-bundle accept/reject rule. The guarded seller used \texttt{base\_attributes/base\_price/optional\_addons} and a chooser that optimized over add-on subsets. Thus policy and transaction-construction scaffold changed together. We reran the same model ladder and profiles with one schema and one chooser in all cells. After parsing, only two booleans---block information and enforce conduct---can change the offer seen by the buyer. The original run contains 180 LLM dialogues in \none{}/\both{}; the controlled $2\times2$ contains 360 LLM dialogues. Fixed-menu rows are deterministic duplicates and excluded from these counts.

\noindent\emph{E2: Repeated-generation forensic.}
From the controlled 14B run, we selected post hoc the four profiles with the largest single-generation \both{}$-$\none{} welfare gains. We then generated each profile-condition three times (24 LLM dialogues; deterministic menu rows are again excluded). We average the three generations within each profile-condition, compute four paired profile effects, and bootstrap profiles. For a finite-sample placebo, we enumerate all $2^4=16$ profile-level sign assignments. We also decompose the balanced profile$\times$condition$\times$replicate table into profile, condition, interaction, and generation-residual sums of squares. Because profiles were selected on their original effects, E2 is an exploratory winner's-curse diagnostic, not a confirmatory treatment estimate.

\noindent\emph{E3: Incentive manipulation check.}
At 14B we compare three seller prompts on three profiles with two generations each: a compliance-oriented seller, the standard ``self-interested'' seller, and a stronger profit-pressure instruction. We inspect seller profit, profit share, rent-extraction questions, and matching behavior. This 18-dialogue smoke test asks whether the textual role manipulation moves the intended construct monotonically; it is not powered to rank prompts.

\noindent\emph{E4: Scripted positive controls.}
We replace the LLM seller with two deterministic policies over all 60 profiles while retaining the same schema, chooser, and scorer. \emph{Profit-maximizing} enumerates offers and chooses the one with highest seller profit (breaking ties by welfare). \emph{Inefficient-bundling stress} forces up to two optional components with $V_{\theta}(o)<C(o)$ into an ungoverned base. Each policy is crossed with the four guardrail cells. These controls ask whether the metrics can detect a welfare-improving guardrail under a seller technology that should produce one.

\subsection{Statistics}

For E1, we compute profile-paired \both{}$-$\none{} welfare differences and 20,000-draw percentile bootstrap intervals, resampling the 30 profiles. For E2, replications are averaged before pairing; raw generation rows never become additional buyer units. We report the exact sign-flip placebo, cell-level generation SD, and a descriptive variance decomposition. The approximate 80\%-power minimum detectable paired effect (MDE) uses $t_{.975,3}+z_{.80}$ times the observed SD of the four profile effects divided by $\sqrt{4}$; its purpose is to expose low precision, not certify prospective power.

\section{Results}

\subsection{R1: The headline effect is scaffold-sensitive}

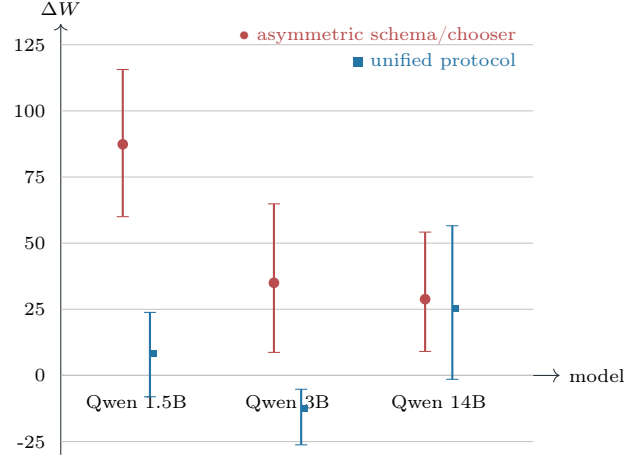
\begin{figure}[t]
\centering
\begin{tikzpicture}[x=1cm,y=1cm,font=\scriptsize]
  \draw[->,draw=ink] (0,1.05) -- (6.6,1.05) node[right]{model};
  \draw[->,draw=ink] (0,0) -- (0,5.7) node[above]{$\Delta W$};
  \foreach \v/\y in {-25/0.175,0/1.05,25/1.925,50/2.8,75/3.675,100/4.55,125/5.425}{
    \draw[draw=gray!45] (0,\y)--(6.25,\y);
    \node[anchor=east] at (-0.08,\y) {\v};
  }
  \foreach \x/\lab in {1/Qwen 1.5B,3/Qwen 3B,5/Qwen 14B}{
    \node[align=center] at (\x,0.68) {\lab};
  }
  \foreach \x/\m/\lo/\hi in {0.82/4.108/3.150/5.097,2.82/2.275/1.355/3.320,4.82/2.059/1.367/2.946}{
    \draw[oldred,thick] (\x,\lo)--(\x,\hi);
    \draw[oldred] (\x-0.08,\lo)--(\x+0.08,\lo) (\x-0.08,\hi)--(\x+0.08,\hi);
    \fill[oldred] (\x,\m) circle (2pt);
  }
  \foreach \x/\m/\lo/\hi in {1.18/1.301/0.768/1.884,3.18/0.565/0.132/0.867,5.18/1.884/0.999/3.030}{
    \draw[controlblue,thick] (\x,\lo)--(\x,\hi);
    \draw[controlblue] (\x-0.08,\lo)--(\x+0.08,\lo) (\x-0.08,\hi)--(\x+0.08,\hi);
    \fill[controlblue] (\x,\m) rectangle +(0.09,0.09);
  }
  \node[anchor=east] at (6.15,5.55) {\textcolor{oldred}{$\bullet$ asymmetric schema/chooser}};
  \node[anchor=east] at (6.15,5.20) {\textcolor{controlblue}{\rule{1.2ex}{1.2ex} unified protocol}};
\end{tikzpicture}
\caption{Profile-paired \both{}$-$\none{} welfare effects (means and 95\% profile-bootstrap intervals, $n=30$). The large 1.5B gain and positive 3B gain do not survive unifying the offer schema and buyer chooser.}
\label{fig:scaffold}
\end{figure}

Figure~\ref{fig:scaffold} shows the central audit result. Under the asymmetric scaffold, mean \both{}$-$\none{} welfare is $+87.4$, $+35.0$, and $+28.8$ for 1.5B, 3B, and 14B. Under the unified schema and chooser, the contrasts are $+7.2$ (95\% CI $[-8.1,23.8]$), $-13.9$ ($[-26.2,-5.2]$), and $+23.8$ ($[-1.5,56.6]$). The estimated effect shrinks by 92\% at 1.5B and reverses sign at 3B. At 14B the point estimate remains positive but is not resolved from zero.

The old result cannot be interpreted as a clean economic effect because C2 is known to fail. The controlled run does not license the opposite claim that guardrails reduce welfare: signs differ by model size, and every cell has only one stochastic generation per profile. The correct aggregate judgment is mixed and under-replicated.

The residual 14B gain is itself a stability warning. Define the $2\times2$ interaction as $I=W_{\both}-W_{\info}-W_{\conduct}+W_{\none}$. It is $-2.5$ at 1.5B and $-0.3$ at 3B, but $+35.8$ at 14B. At 14B both single-rule main effects are negative ($-3.5$ and $-8.4$), so an additive model predicts welfare $93.5$; the observed \both{} welfare is $129.3$. This isolated super-additivity accounts for the surviving positive contrast but appears at only one model size with one generation per profile-condition. The aggregate channel is primarily completion: \both{} raises completion from $.77$ to $.90$, while accepted-trade welfare rises only from $137.5$ to $143.6$. Holding accepted-trade welfare at the \none{} level, the completion change accounts for $18.3$ of the $23.8$ aggregate-welfare units, or 77\%. We therefore treat the interaction as a post-hoc diagnostic, not evidence of complementarity.

\subsection{R2: Single generations overstate selected effects}

\begin{figure}[t]
\centering
\begin{tikzpicture}[x=1cm,y=1cm,font=\scriptsize]
  \draw[->,draw=ink] (0,1.25)--(6.2,1.25);
  \draw[->,draw=ink] (0,0)--(0,5.0) node[above]{$\Delta W$};
  \foreach \v/\y in {-100/0.2,0/1.25,100/2.3,200/3.35,300/4.4}{
    \draw[draw=gray!45] (0,\y)--(4.25,\y);
    \node[anchor=east] at (-0.08,\y) {\v};
  }
  \foreach \x/\lab in {0.65/B0209,1.65/B0222,2.65/B0225,3.65/B0227}{\node at (\x,0.90) {\lab};}
  \foreach \x/\old/\new in {0.65/3.886/2.129,1.65/3.256/0.533,2.65/4.085/1.250,3.65/3.392/2.668}{
    \draw[gray!60,thick] (\x-0.10,\old)--(\x+0.10,\new);
    \fill[oldred] (\x-0.10,\old) circle (2pt);
    \fill[controlblue] (\x+0.10,\new) rectangle +(0.08,0.08);
  }
  \node[anchor=west] at (4.55,4.35) {\textcolor{oldred}{$\bullet$ single}};
  \node[anchor=west] at (4.55,4.00) {\textcolor{controlblue}{\rule{1.2ex}{1.2ex} mean of $k=3$}};
  \node[draw=ink,rounded corners=1.5pt,fill=softgray,align=left,text width=2.9cm,anchor=north west] at (4.45,3.65) {
    single mean: $+229$\\
    $k=3$ mean: $+37.6$\\
    95\% CI: $[-34.2,109.3]$\\
    mean rep SD: $47.8$\\
    exact placebo: $p=.50$
  };
\end{tikzpicture}
\caption{Winner's-curse probe. Profiles were selected post hoc for the largest positive 14B single-generation effects. Repetition changes both magnitude and sign.}
\label{fig:replication}
\end{figure}
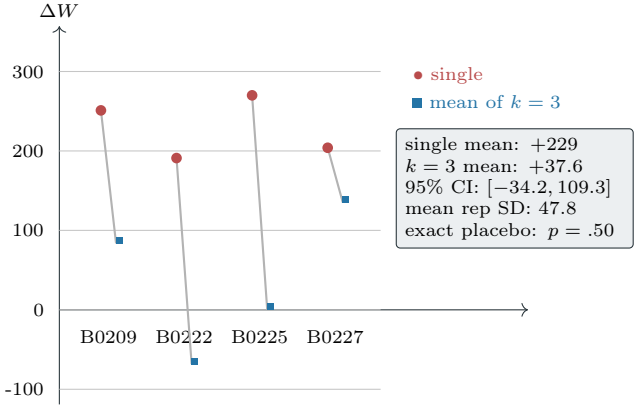

The four largest 14B single-generation effects were $+251$, $+191$, $+270$, and $+204$ welfare units. Their mean was $+229$. With three new generations per cell, the profile effects become $+83.7$, $-68.3$, $0$, and $+135$, with mean $+37.6$ and profile-bootstrap interval $[-34.2,109.3]$ (Figure~\ref{fig:replication}). Mean within-cell generation SD is $47.8$. The approximate 80\%-power MDE is $180.9$, nearly five times the repeated point estimate.

The exact profile-level sign-flip distribution gives a two-sided $p=0.50$ (one-sided $p=0.25$). In the descriptive variance decomposition, generation residuals account for 49.9\% of sum-of-squares variation, profile effects 24.1\%, profile-by-condition interaction 21.1\%, and condition 4.9\%. This occurs despite temperature $0.2$: low-temperature decoding did not make one rollout a stable measurement. These percentages apply only to the selected four-profile probe, but they show why the initial extreme effects cannot be treated as stable profiles.

\subsection{R3: The assumed strategic seller is not validated}

Table~\ref{tab:incentive} reports the seller manipulation check. The profit-pressure prompt increases rent-extraction questions relative to the standard prompt ($0.83$ vs. $0.17$ per dialogue), but it produces much less seller profit ($33.8$ vs. $57.5$) and a lower profit share ($0.37$ vs. $0.56$). The compliance prompt yields the least profit, as expected, but the intended middle-to-high ordering fails. This is not evidence that stronger prompts generally reduce profit; $n=3$ profiles is too small. It is evidence that the current prompt manipulation has not established a monotone, controllable profit-maximization construct. C1 is therefore \inconclusive{}, not a known violation.

\begin{table}[t]
\centering
\scriptsize
\setlength{\tabcolsep}{2.0pt}
\begin{tabular}{lrrrr}
\toprule
Seller prompt & Profit & Share & Rent probes & Welfare \\
\midrule
Compliance & 25.8 & .29 & .17 & 92.2 \\
Standard self-interest & 57.5 & .56 & .17 & 103.2 \\
Profit pressure & 33.8 & .37 & .83 & 94.7 \\
\bottomrule
\end{tabular}
\caption{Seller-incentive smoke test (Qwen2.5-14B, 3 profiles, 2 generations each). Stronger profit language changes probing but not profit monotonically.}
\label{tab:incentive}
\end{table}

\subsection{R4: Positive controls locate the missing mechanism}

The scripted controls separate metric failure from construct absence (Table~\ref{tab:controls}). The profit-maximizing seller extracts all buyer surplus in \none{} but selects a first-best bundle, yielding mean welfare $169.1$ and 100\% of first best. This is the textbook complete (first-degree) price-discrimination benchmark: in a frictionless environment the seller captures surplus while preserving the efficient allocation \cite{varian1989pricediscrimination}. Adding both guardrails transfers $44.0$ to buyers, removes $68.5$ from sellers, and \emph{reduces} welfare by $24.5$. Thus seller self-interest and buyer extraction alone do not imply a welfare-improving guardrail.

The inefficient-bundling stress seller establishes the positive control. It forces components whose buyer value is below seller cost when conduct is ungoverned. Under this technology, \both{} raises welfare by $18.7$ and first-best attainment by 11.1 percentage points. Yet the same intervention still lowers seller profit by $25.3$ while increasing buyer surplus by $44.0$. Reporting buyer surplus alone would make both scripted worlds look equally successful; joint accounting reveals opposite welfare signs.

\begin{table}[t]
\centering
\small
\setlength{\tabcolsep}{4.0pt}
\begin{tabularx}{\columnwidth}{@{}Xrrrr@{}}
\toprule
Scripted seller & $\Delta$BS & $\Delta$SP & $\Delta W$ & $\Delta$FB pp \\
\midrule
Profit maximizing & $+44.0$ & $-68.5$ & $-24.5$ & $-15.4$ \\
Inefficient-bundling stress & $+44.0$ & $-25.3$ & $+18.7$ & $+11.1$ \\
\bottomrule
\end{tabularx}
\caption{Positive controls, \both{}$-$\none{}, 60 assigned buyer types. FB pp denotes percentage-point change in profile-normalized first-best welfare. The identical $\Delta$BS values are not a copy error: under \both{}, the two scripted controls induce the same constrained outcomes. The same buyer-surplus gain masks opposite welfare effects.}
\label{tab:controls}
\end{table}

\subsection{R5: The contract abstains selectively}

Table~\ref{tab:verdict} applies C1--C4 rather than leaving the contract as advice for other studies. The original headline is \invalid{} because its treatment changes the transaction scaffold. The controlled estimand is meaningful, but the available LLM evidence remains \inconclusive{} because incentive validity is unverified and generation stability is unresolved. C4 passes: the accounting identity and scripted controls distinguish transfer from welfare creation. Accordingly, neither ``guardrails work'' nor ``guardrails fail'' is licensed. This abstention concerns the causal policy estimand, not the audit findings: scaffold sensitivity, the variance decomposition, and the scripted-control contrast are themselves measured results.

\begin{table}[H]
\centering
\scriptsize
\setlength{\tabcolsep}{2.0pt}
\begin{tabularx}{\columnwidth}{@{}p{2.0cm}>{\raggedright\arraybackslash}p{2.3cm}X@{}}
\toprule
Contract item & Verdict & Evidence in this audit \\
\midrule
C1 Incentives & \inconclusive{} & Non-monotone smoke test; only 3 profiles and 2 generations. \\
C2 Protocol & Original: \invalid{}\newline Controlled: Pass & Original schema and chooser differ; E1 removes both differences. \\
C3 Stability & \inconclusive{} & MDE $180.9 \gg 37.6$; exact sign-flip $p=.50$; 49.9\% residual variation. \\
C4 Accounting & Pass & $\Delta$BS, $\Delta$SP, $\Delta W$, completion, and first-best are jointly reported. \\
\midrule
Overall & No causal policy claim; descriptive reporting licensed & C2 blocks the original; C1 and C3 block the causal interpretation of the controlled study. \\
\bottomrule
\end{tabularx}
\caption{The evaluation contract applied to the present study. A failed gate determines the admissible interpretation; passing one gate does not compensate for another.}
\label{tab:verdict}
\end{table}

\section{Implications for Agent Evaluation}

\noindent\emph{Validate roles as manipulations, not prose.}
An agent called ``self-interested'' should not be presumed to implement a utility function. Before testing governance, researchers should specify a behavioral implication of the role---for example, monotone response to a profit coefficient, revealed-preference consistency, or regret against a scripted optimum---and test it. If the manipulation fails, the study may still report descriptive LLM behavior, but not a policy effect on strategic sellers.

\noindent\emph{Hold the transaction constructor fixed.}
Agent evaluations often bundle a policy with templates, parsers, normalization, retries, and decision logic. Those components can dominate the intervention. Our strongest initial result came from exactly such a bundle. A defensible ablation uses one schema and one chooser, then applies treatment flags after parsing. When this is impossible, scaffold changes should be named as part of the intervention rather than attributed to a thin guardrail.

\noindent\emph{Treat generations as nested measurements.}
Repeated generations estimate within-cell stochasticity; they do not create new buyer types. The simplest correct workflow is to average $k$ generations within profile-condition, form paired profile effects, and resample profiles. A mixed model is another option. Pairing ``replicate 1'' across two independently sampled conditions and treating $n\times k$ rows as independent narrows uncertainty without creating information.

\noindent\emph{Make abstention an evaluation output.}
Evaluation reports typically force an answer even when their own assumptions fail. Our contract makes two forms of abstention explicit. \invalid{} means a known design violation prevents the named interpretation, as in the asymmetric scaffold. \inconclusive{} means the estimand is meaningful but precision or coverage is insufficient, as in the repeated four-profile probe. Neither should be translated to a substantive null.

The accompanying artifact turns these gates into a nine-item checklist covering roles, schemas, trace retention, replication, accounting, and controls. It also provides the analysis script, generated tables, prompts, and run manifests at \url{https://anonymous.4open.science/r/a2a-evaluation-artifact-staging-2F25/}.

\noindent\emph{Separate creation from transfer.}
For accepted trades, price cancels from welfare. Consequently, a privacy or disclosure rule may improve buyer surplus while reducing seller profit one-for-one. A platform may care about that distribution, but it is not total-welfare creation. Every agent-market report should show completion, buyer surplus, seller profit, and welfare together and state which outcome is normative.

\noindent\emph{Broader impact.}
More reliable evaluation can prevent unsupported marketplace policies and make null or negative evidence visible. The same audit language can create false assurance if its contract omits privacy, fairness, or deployment harms; retaining buyer profiles and transcripts for audit can also expose sensitive preferences. We mitigate these risks here by using synthetic data, releasing no model weights or personal data, separating descriptive from causal claims, and requiring \invalid{} or \inconclusive{} rather than a forced policy verdict. Deployment would still require domain-specific privacy, fairness, and governance review.

\section{Limitations and Scope}

This audit is intentionally narrower than the claims it warns against. First, all LLM results use one instruction-tuned model family, three sizes, 4-bit quantization, and self-play. Using the same model for buyer and seller may dampen adversarial behavior through shared priors or a common instruction-following style; cross-family pairing with role prompts held fixed is the cheapest direct test. We have not shown that frontier or cross-family sellers fail the same incentive check. Second, the repeated-generation analysis covers four profiles selected after observing extreme effects. It diagnoses instability and winner's curse but cannot estimate a population treatment effect. A powered confirmatory run would require pre-specified profiles and at least three generations per profile-condition.

Third, the hotel profiles are synthetic. Their outside-option distribution includes 43 negative utilities among 60 profiles, so most participation constraints weakly bind; with only two question rounds, the environment may not afford the extraction dynamics that the information rule is meant to alter. A null \info{} effect would therefore be confounded by the opportunity structure, not evidence that information protection is irrelevant. The optimized fixed menu also has a training-distribution advantage over zero-shot agents, so we do not use agent-versus-menu performance as a headline. Fourth, scripted sellers are instruments, not realistic behavioral models; they prove that the scorer responds under explicit technologies, not that real sellers behave that way. Finally, our contract is a minimum gate, not a complete theory of external validity, platform equilibrium, or human welfare.

\section{Conclusion}

An LLM market simulation can produce precise-looking prices and welfare while failing to instantiate the seller behavior or isolated policy intervention needed to interpret them. In our case study, a strong guardrail result was largely scaffold-sensitive, its most favorable profiles were generation-unstable, and the seller-incentive manipulation did not move profit monotonically. Scripted controls then showed that guardrails create welfare only under a specific welfare-destroying seller technology; otherwise they may merely redistribute or reduce it. Agentic evaluation should therefore verify incentives, isolate protocol changes, nest replications correctly, and report complete welfare accounting before making a policy claim. When those checks fail, \invalid{} and \inconclusive{} are results, not disclaimers.

\bibliographystyle{plain}
\bibliography{references}

\end{document}